# Increasing Adverse Drug Events extraction robustness on social media: case study on negation and speculation

## Authors

Simone Scaboro[1], Beatrice Portelli [1,2], Emmanuele Chersoni [3], Enrico Santus [4], Giuseppe Serra [1]

[1] University of Udine - Department of Mathematics, Computer Science and Physics, via delle Scienze 206, Udine 33100, IT

[2] Università degli Studi di Napoli Federico II, Corso Umberto I 40, Napoli, 80138 IT.

[3] The Hong Kong Polytechnic University - Department of Chinese and Bilingual Studies, Hung Hom, HK.

[4] Decision Science and Advanced Analytics for MAPV & RA, Bayer, Bayer Pharmaceuticals, Whippany, US.

## Abstract

In the last decade, an increasing number of users have started reporting Adverse Drug Events (ADE) on social media platforms, blogs, and health forums. Given the large volume of reports, pharmacovigilance has focused on ways to use Natural Language Processing (NLP) techniques to rapidly examine these large collections of text, detecting mentions of drug-related adverse reactions to trigger medical investigations. However, despite the growing interest in the task and the advances in NLP, the robustness of these models in face of linguistic phenomena such as negations and speculations is an open research question. Negations and speculations are pervasive phenomena in natural language, and can severely hamper the ability of an automated system to discriminate between factual and non-factual statements in text. In this paper we take into consideration four state-of-the-art systems for ADE detection on social media texts. We introduce SNAX, a benchmark to test their performance against samples containing negated and speculated ADEs, showing their fragility against these phenomena. We then introduce two possible strategies to increase the robustness of these models, showing that both of them bring significant increases in performance, lowering the number of spurious entities predicted by the models by 60% for negation and 80% for speculations.

## Impact Statement

The work studies the shortcomings of the current models for Adverse Drug Event (ADE) detection models for social media texts against linguistic phenomena, such as negation and speculation, which often undermine the capacity of information extraction systems to identify factual statements. Monitoring social media can allow medical and pharmaceutical agencies to collect crowd signals related to the products in the market, and current research is moving in this direction. However, social media data are noisy and may lead to incorrect interpretation without properly-designed algorithms. In our work, we introduce two viable strategies to improve the robustness of current ADE extraction methods against negation and speculation, and demonstrate their effectiveness. We also provide to future researchers a new benchmark dataset, which contains, for the first time, samples of social media texts annotated for the presence of negated and speculated ADEs, and all the related code.

# Introduction

Every year, several drugs are developed and tested, and are finally approved by regulators such as the Food and Drug Administration (FDA) or the European Medical Agency (EMA). The safety of these substances is tested in medical trials. However, it is impossible to foresee all possible collateral effects and adverse reactions that might occur in a very large population: the drugs will be administered to patients of different ages, ethnicities, and medical conditions, for prolonged periods, and interacting with other substances. Therefore Pharmacovigilance (PV) has the task to monitor the drugs after they entered the market, detecting and cataloging all Adverse Drug Event (ADE) reports.

Traditionally, ADE reports are collected through the communication between patient, healthcare provider and PV authorities. However, previous studies have shown that over 90% of the ADEs complained by patients are not forwarded to local or regional PV systems.[1] This causes an underestimation of the incidence of harmful side effects, and slows down the discovery of new and rare phenomena, especially the ones related to under-represented categories of patients.

At the same time, an increasing number of Internet users started sharing their health details and experiences on the Internet, thanks to forums and microblogging platforms, such as Facebook and Twitter. Exploring social media texts has therefore become increasingly important in the field of PV,[2-3] as they can potentially provide an updated stream of user feedback on medications, drugs and procedures, through information mining and gathering of crowd signals. The main drawback of this source of data is its inherent noisiness. Social media texts contain colloquial language, typos, slang, metaphors, and non-standard syntactic constructions. They also present the additional challenge of containing implicit information, because of the short format, and lack the structured form of official ADE reports.[4]

In recent years, the Natural Language Processing (NLP) community has dedicated a consistent effort in developing robust methods for mining user-level medical information from social media texts. This also led to the creation of several dedicated shared tasks series on ADE detection, such as the "Social Media Mining for Health Application Workshop and Shared Task" (SMM4H).[5-9]

The models developed for these tasks have seen great advancements in the last few years, thanks to new neural architectures such as those based on pre-trained Transformers.[10-11] However, their robustness against the possible presence of both negations and speculations, has never been investigated. Generally speaking, handling negation and speculation scopes effectively is a known problem in NLP when applied to formal texts,[12-13] but it has never been studied in the context of ADE extraction from informal texts.

Negation and speculation detection are essential to recognize if a post is claiming or denying a cause-effect link between a drug and an ADE, otherwise this could lead to the extraction of non factual information. For example, if we consider the tweet: "That's correct! Metoprolol is NOT known to cause hypokalemia", ADE extraction models might incorrectly identify "hypokalemia" as an ADE of Metoprolol. The consequences of erroneous ADE extractions, such as the one mentioned above, could mislead the experts on possible drug side effects, and create additional noise that steals the spotlight from the reports that contain actually useful information.
When dealing with social listening on large amount of data, it is of paramount importance to minimize the amount of noise produced by the AI extraction systems, in order to reduce the burden for the human experts who are going to analyze it. This is true, even when it comes with the cost of missing some true ADE mentions.
This might seem highly undesirable at first, as usually in medical tasks it preferred to cause a false alarm (i.e., producing False Positives) than miss an actual ADE mention (i.e., generating False Negatives). However, losing some true mentions is not as alarming in the context of gathering crowd signals. Even if we lose a mention for a specific Adverse Event (e.g., "no sleep"), we are likely to find several other reports of the same problem expressed in different terms, that the system is able to detect (e.g., "lack of sleep" or "sleepless night"). These kinds of ADEs will be eventually picked up by the final system thanks to data aggregation, and False Negatives are therefore less detrimental than False Positives.
Of course, this is not true in the case of diagnosing a single patient (e.g., ADE extraction from Electronic Health Records): in this situation, a high FN rate is unacceptable, as each episode is processed on its own.

The objective of this work is to investigate the robustness of current ADE extraction methods for social media texts in presence of negations and speculations, to propose strategies to tackle these issues and to create more reliable ADE extraction models.
To analyze the robustness of current, ADE extraction system, we introduce the SNAX (Speculations and Negations for ADE eXtraction) benchmark, an extension of NADE,[14] with the addition of speculated samples. First, we analyze several state-of-the-art models, testing their performance against the new challenging samples. We also introduce two effective strategies to increase the robustness of current models: data augmentation, augmenting the training set with artificially negated and speculated samples; model combination, adding a negation/speculation detection module in a pipeline fashion to exclude the non-factual ADEs.
We show that the first strategy can decrease the number of excess predictions by 60% for both negated and speculated samples, while the second one can remove 50% of the excess predictions in case of negations, and up to 80% in the case of speculations. The two strategies can also be combined, maintaining the effects of both.
To further the research in this field and allow researchers to test the robustness of their own ADE extraction systems, all the materials produced in this work are publicly available on GitHub at https://github.com/AilabUdineGit/SNAX.

## Related Work

### ADE detection in Social Media Texts

The extraction of ADE mentions from social media texts is a task which started receiving growing attention in the last few years. This was prompted by the increasing number of users who discuss their drug-related experiences online, on platforms such as Twitter. The first studies which proposed machine learning approaches for this task used either traditional feature engineering or word embeddings-based solutions.[3,15-16] The problem gained visibility with the introduction of the SMM4H Workshop and the shared tasks co-located with it. Currently, the SMM4H datasets are the largest collection of tweets tagged for the presence of ADE, and they are updated yearly. Methods based on neural networks became a more common choice. In the SMM4H'19 competition, transformers-based architectures such as BERT and BioBERT were the building blocks of the top performing systems[17-19] and later works have investigated the characteristics of the most effective pretrained BERT-like models for this task.[20] However, recent studies[21] showed that these NLP models are not capable of properly understanding some pervasive linguistic phenomena, among which are negations and speculations.

On the other hand, in most of the existing ADE detection datasets,[22-23] samples are only marked for the presence or absence of an ADE. Samples that negate or speculate an ADE are under-represented and conflated with the "no ADE" class. This is why we worked towards the creation of a more comprehensive benchmark dataset.

In our preliminary version of this work[14] we started analyzing the effect of negations on the accuracy of ADE extraction models. Here we extend the analysis to also include the speculation phenomenon, which is as widespread as the negation one, and less addressed in the literature. We extend the previously released dataset with new samples containing speculated ADE; we also introduce custom speculation detection modules, based on regular expressions and neural network, to detect speculated samples; finally, we analyze the effect of various strategies to improve model robustness on a series of well-performing ADE extraction models, including new ones introduced in the latest SMM4H'20 Shared Task.

## Negation and Speculation Detection in Biomedical Texts

We introduce a brief overview on general methods for negation and speculation detection, as some of these techniques will be used in the strategies introduced in this work. In particular, this research field has been dominated by techniques based on dictionaries and regular expressions. In the case of negation detection, NegEx[24] was one of the earliest and most popular systems, and was introduced to identify negations in English clinical notes. Following developments of this work included extensions such as the identification of the experiencer (i.e., the person who is experiencing the event) and of the temporal status of symptoms.[25] Efforts have also adapted the systems to European languages other than English.[26]

Regular expressions were later replaced by machine learning approaches, especially after the publication of the BioScope corpus,[27] a large dataset of biomedical documents annotated for the presence of negation and speculation scopes. Most of the proposed strategies consisted of a two-step methodology: first, a classifier to detect negation/speculation cues in the sentence (i.e., words that signal the presence of these phenomena, such as "not", "never" or "unproven"), second, a classifier to determine which part of the sentence falls into the scope of a cue word (i.e., which concept is negated or speculated).[28-31]

The latest approaches are based on neural networks, in particular CNNs[32] and BiLSTMs,[33-35] which have shown capabilities to transfer knowledge across languages and domains. Since the introduction of BERT in the field of NLP, researchers have also introduced BERT-based models for negation detection,[36] speculation detection,[37] and others which are also aided with multitask learning architectures.[38]

### Adversarial Benchmarking

Another relevant line of research for our work is related to adversarial benchmarking.[39] Adversarial benchmarking became popular as a response to trends in machine learning research emphasizing that models are surpassing estimates of human performances. Researchers pointed out that the problem arises from the current inadequacy of the current benchmarks, which are affected by social and statistical biases that make them artificially easy, and not adequately reflecting the complexity and adaptability of human linguistic competence.

Adversarial benchmarking consists in collecting data against state-of-the-art models in the loop, over multiple rounds of annotation and generation of text: human annotators are trained to generate new training examples including features that make them challenging for automatic systems. Using this new approach to benchmarking, several new evaluation datasets for classical NLP tasks have been created, e.g. sentiment analysis,[40] question answering,[41] and hate speech detection.[42] In our case, we chose to focus on negation and speculation cues for the creation of the adversarial examples because those phenomena have been proven to fool state-of-the-art Transformer models.[14]

However, differently from the common practice in adversarial benchmarking, in this paper we create a single set of examples of negations and speculations instead of running multiple annotation rounds, given our limited availability of annotators. We leave the task of a more extensive annotation work and the creation of a bigger adversarial benchmark for ADE detection to future work.

# Materials and Methods

## SNAX Dataset

In this work, we proposed a dataset for ADE detection on social media texts, which includes negations and speculations. In particular, the datasets is composed by tweets, which can be subdivided in the following four classes:

- samples containing ADEs ($\mathbb{A}$);
- samples not containing ADEs ($\mathbb{X}$);
- samples that explicitly negate the presence of an ADE ($\mathbb{N}$);
- samples that speculate or question the presence of an ADE ($\mathbb{S}$).

We collected $\mathbb{A}$ and $\mathbb{X}$ samples from the dataset of the SMM4H ADE extraction challenge,[8] which is the largest and most up-to-date collection of tweets annotated for the presence of ADEs.

As regards the $\mathbb{N}$ and $\mathbb{S}$ samples, we started by recovering from the SMM4H binary classification datasets.[8-9] Due to the way the SMM4H dataset was created, negated and speculated samples were scarcely present, so we decided to create additional artificial $\mathbb{N}$ and $\mathbb{S}$ samples. We selected a number of tweets containing ADEs and manually altered them to negate or speculate the Adverse Event. We decided to use all artificial samples as part of the training data, and leave all real samples in the test set. In this way, the models will only be tested on real data.

The process to recover real $\mathbb{N}$/$\mathbb{S}$ samples and generate artificial ones was carried out by four volunteers with a high level of proficiency in English. More specifically, the volunteers were: two graduate students (Master's degree in Computer Science and Artificial Intelligence) and two Ph.D. in Natural Language Processing. All of them have a minimum English level of C1 (https://www.efset.org/cefr/c1/).

**Recovery of Real ℕ/𝕊 Samples**

The SMM4H classification dataset contains a great number of tweets that mention drug names but contain no ADE. Some of them might belong to the ℕ or 𝕊 class, but they are far too many to check manually for their presence. We used a list of possible negation and speculation cues to pre-filter the data and ease the work of the human annotators. This produced approximately 4000 samples, which were manually analyzed by the human annotators to assess whether the negation/speculation cue referred to an ADE.

The following are examples of tweets that contain a negation/speculation cue (bold), but do not include an actual ADE, and have not been selected to be part of the augmented dataset:

- But I'm **not** on adderall and I am feasting.
- I've seen so much Tamiflu these past couple of days I'm **not** even surprised I'm shivering and experiencing aches right now. *sigh
- After that game, Doc emrick **may** need a #lozenge
- I have **no evidence** to back up that bold assertion other than the sudden proliferation of uloric commercials

The following are insead examples of tweets that were selected to be part of the augmented data, as they actually negate or speculate an ADE (bold):

- This #HUMIRA shot has me feeling like a normal human... **No pain no inflammation** no nothinggggh #RAproblems
- @UKingsbrook That's correct! Metoprolol is **NOT known to cause hypokalemia**.
- that medicine humira **seems like it has more see effects** than reliefs
- #FDA is investigating a **possible link** between the type 2 #diabetes drug saxagliptin (Onglyza) and **heart failure**, the agency has announced

The selected tweets were reviewed by all annotators, and we only kept the ones for which they were in agreement. As a result, we obtained two new sets of real ℕ and 𝕊 tweets.

**Generation of ℕ/𝕊 Artificial Samples**

We used 260 𝔸 samples from the SMM4H'19 extraction datasets to modify them and generate artificial ℕ and 𝕊 samples. The generation process was carried out by the same four annotators. Each of them was given part of the 260 tweets and was instructed to alter them to generate a new version of the sample, that either negated or speculated the presence of the ADE. Conservative edits were encouraged, but if it was not possible to negate the meaning of the tweet just by adding cue words, the annotators were allowed to perform more edits in the sentence and use longer expressions.

Each volunteer was asked to review the tweets generated by the other participants and to propose modifications, if the tweets looked ambiguous, unnatural, or failed to negate/speculate the ADE. If no agreement was reached, the augmented tweet was discarded. At the end of the process, the annotators generated 251 artificial ℕ tweets and 227 artificially 𝕊 tweets. We can observe that the speculation process was more challenging, as the annotators failed to reach consensus on 33 speculated samples, and only 9 negated samples.

Here is an example of an original tweet and its negated and speculated versions (highlighting the phrases added to alter the meaning of the tweet):

- #restlesslegs #quetiapine
    - **no** #restlesslegs for me on #quetiapine
    - really **possible** #restlesslegs with #quetiapine**?**
- I'm on citalapram and 600mg quetiapine - i'm not sleeping!
    - I'm on citalapram and 600mg quetiapine - **and there's no way** I'm not sleeping, **gnight**!
    - I'm on citalapram and 600mg quetiapine - **probably** i'm not going to sleep!

## Analyzed models

To increase the robustness of current ADE extraction systems, we introduce and test two general strategies, which can be applied independently from the chosen model: data augmentation and model combination.

Figure 1. Schema and example of use of the data augmentation strategy. Usually, an ADE extraction module is only trained on samples of two classes 𝔸 and 𝕏 (top). With this strategy, we also add negated (ℕ) and/or speculated (𝕊) samples to the training set, increasing the model’s knowledge on these phenomena.

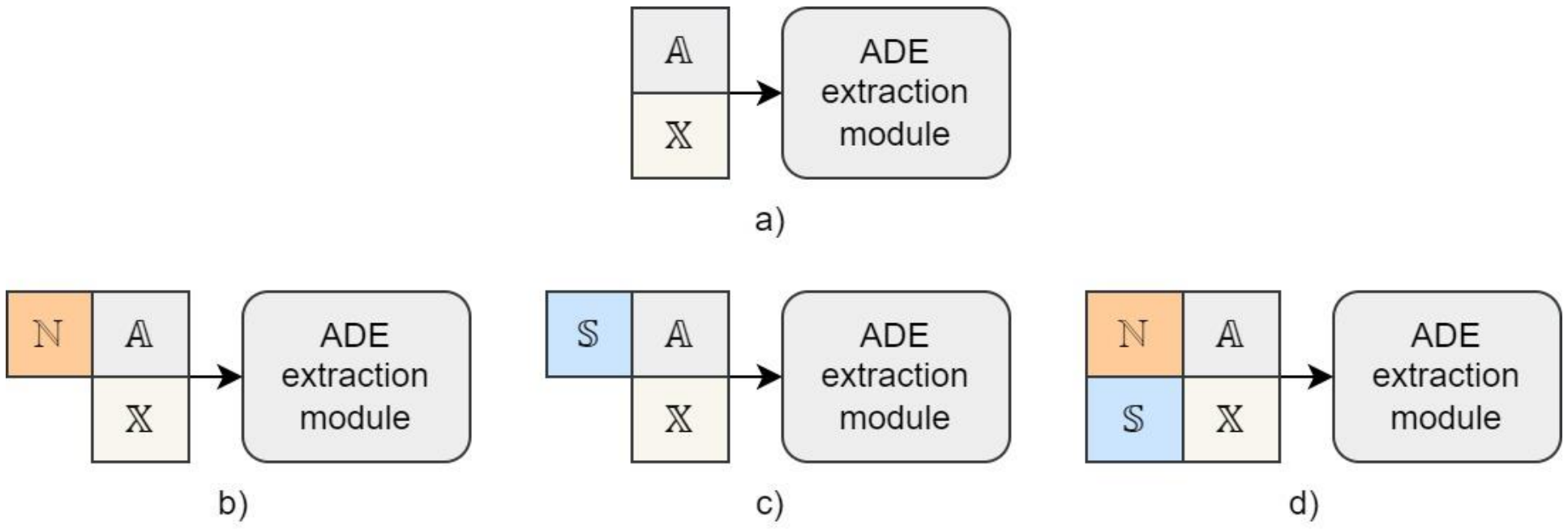

The first one consists in training the models on a dataset that includes examples of the linguistic phenomena that we want to avoid (i.e., negation and speculation). ADE extraction models are usually trained only on 𝔸 and 𝕏 samples (see Figure 1 a), while the data augmentation strategy introduces ℕ training samples (Figure 1 b), 𝕊 samples (Figure 1 c) or both (Figure 1 d), so that the model can learn how to deal with them.

The second strategy to increase the system’s robustness is model combination, that is adding a negation and/or speculation detection module in a pipeline fashion to exclude ADEs which are negated or speculated. The intuition behind this idea is that, given a tweet (e.g., “Metroprolol is NOT known to cause hypokalemia”, see Figure 2 a), an ADE extraction module will predict a set of possible ADEs (e.g., “hypokalemia”), while the negation extraction module will predict a separate set of possible negated scopes (e.g., “NOT known to cause hypokalemia”). We can then check if any of the ADEs intersect the negations. If they do, we discard them, assuming that the tweet is negating the presence of the ADE. The same can be done with a speculation detection module (see Figure 2 b), and multiple different modules can be applied to the same tweet (see Figure 2 c).

Figure 2. Schema and example of use of the combined models. a) shows the effect of combining an ADE extraction module with a negation detection module. b) shows the combination of an ADE extraction module with a speculation detection one. c) illustrates the effect of combining all three modules.

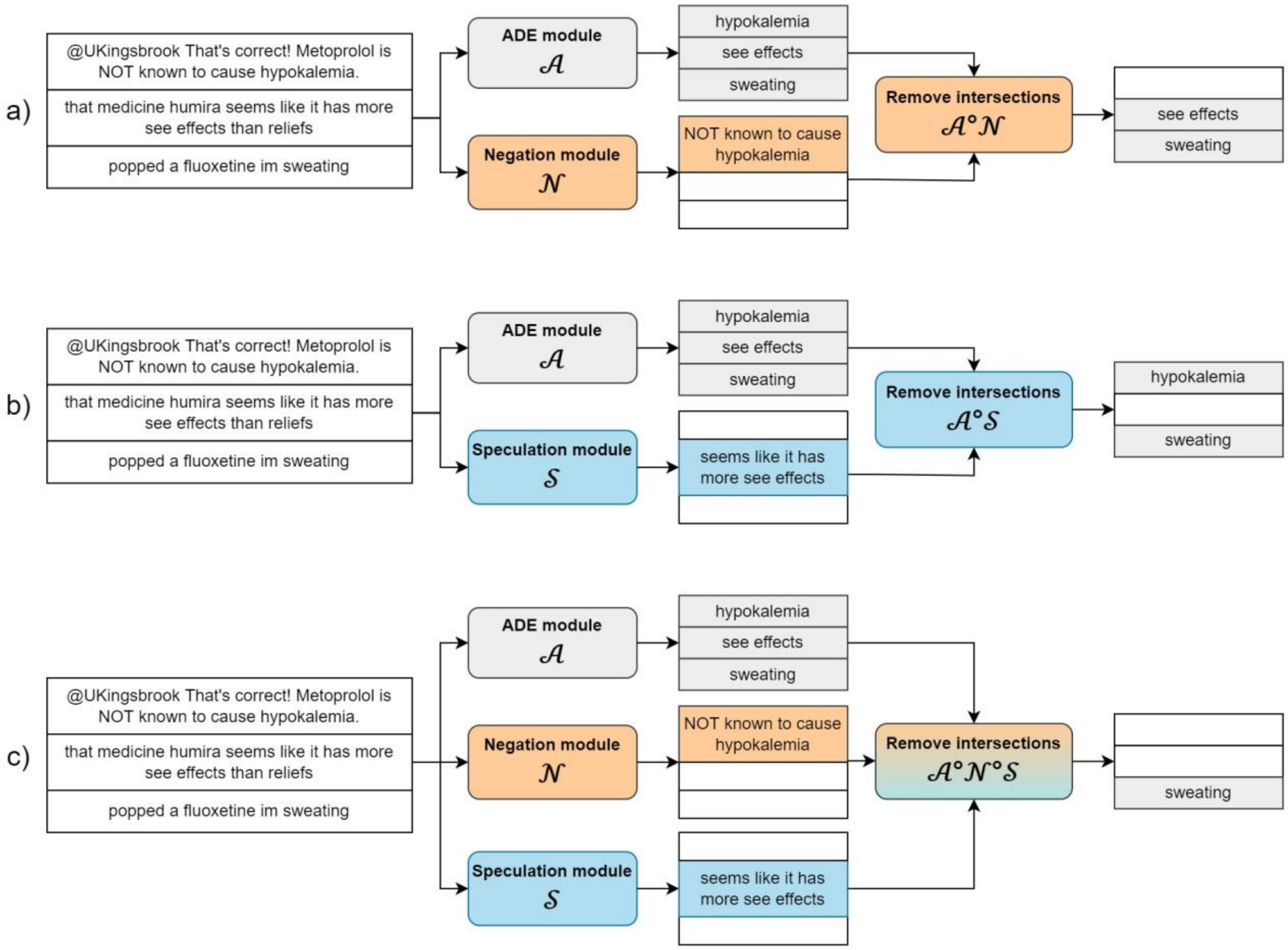

Following is a more formal description of the process. Let us consider a text t and an ADE extraction model $\mathcal{A}$. Given t, $\mathcal{A}$ outputs a set of substrings of t which it considers to be ADE mentions:

$$\mathcal{A}(t) = \{a_1, a_2, \ldots, a_n\}$$

Similarly, a negation extraction model $\mathcal{N}$ and a speculation extraction model $\mathcal{S}$ take a text as input and output a set of its substrings, which they consider to be entities within a negation/speculation scope:

$$\mathcal{N}(t) \;=\; \{n_1, n_2, \ldots, n_j\}$$
$$\mathcal{S}(t) \;=\; \{s_1, s_2, \ldots, s_k\}$$

A combined model is obtained by discarding all the ADE spans $\{a_i\} \in \mathcal{A}(t)$ that overlap with a negation/speculation span:

$$\mathcal{A}°\mathcal{N}(t) \;=\; \{a_i \in \mathcal{A}(t) \mid \forall n_j \in \mathcal{N}(t)\ (a_i \cap n_j = \emptyset)\}$$
$$\mathcal{A}°\mathcal{S}(t) \;=\; \{a_i \in \mathcal{A}(t) \mid \forall s_j \in \mathcal{S}(t)\ (a_i \cap s_j = \emptyset)\}$$
$$\mathcal{A}°\mathcal{N}°\mathcal{S}(t) = \mathcal{A}°\mathcal{N}(t) \cap \mathcal{A}°\mathcal{S}(t)$$

## ADE Extraction Models

We choose three BERT-based models that showed high performance on the SMM4H'19 extraction dataset,[20,43] and are currently at the top of the SMM4H'19 extraction ADE extraction leaderboard:

BERT,[11] SpanBERT[44] and PubMedBERT[45]. We also include EnDR-BERT,[46] which was one of the best performing models in the SMM4H'20 extraction task.
We used the same hyperparameters reported by Portelli et al.[20] The models are fine-tuned to predict a label for each input word, marking the presence of ADEs with the BIO annotation schema (Begin, Inside, Outside). The BIO annotation schema is commonly used for Named Entity recognition tasks and consists in assigning a B label to the first word of the entity (i.e., an ADE), an I label to all the following ones and an O label to all other words (which are not part of any entity). Figure 3 shows an example of BIO tagging for an ADE sample, and how it is processed by the BERT-based models.

Figure 3. Example of BIO (Begin-Inside-Outside) tagging for an ADE sample and schema of the inputs and outputs of the BERT-based models used for the experiments. The model takes as input a sentence, and outputs a BIO label for each of its words. The B and I labels identify the predicted ADE spans.

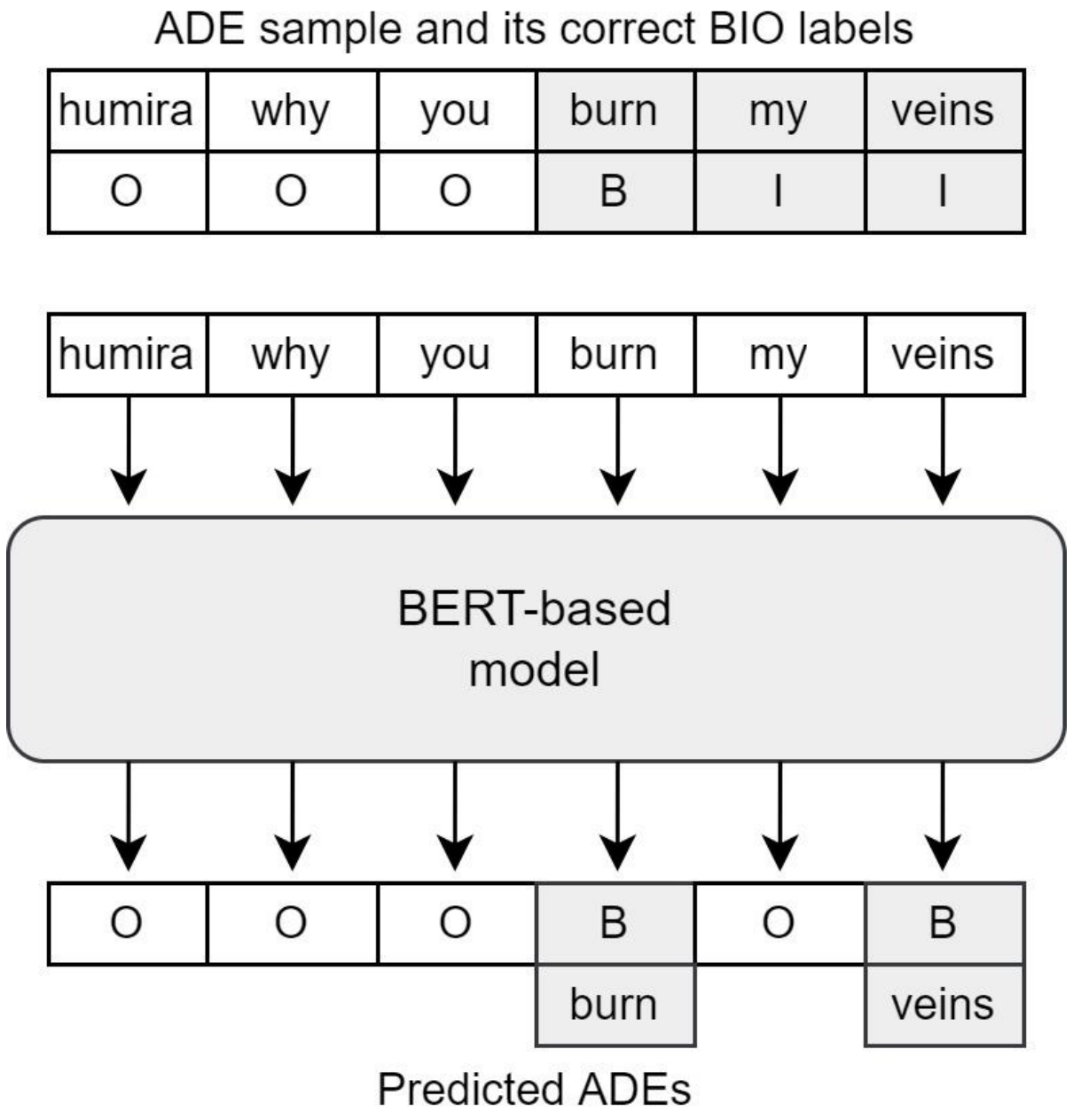

## Negation Detection Models

We evaluate two negation detection models. First, we use NegEx[24], as it is a simple yet effective algorithm based on regular expressions, and it is widely used. We also implement NegB, a negation detection module based on attentive neural networks. Starting from a pretrained BERT model, we finetune it for token classification, training it on the BioScope[27] dataset. We selected 3190 sentences (2800 of which with a negation scope) and trained the model for scope detection (10 epochs, learning rate 0.0001).

## Speculation Detection Models

To detect speculations, we implement two modules, one based on regular expressions, and one based on pretrained BERT models, similarly to what described above. We developed SpecEx, a variation of the NegEx algorithm specialized for speculation detection. In order to do that, we extracted all the speculative cues present in the BioScope corpus, divided them into categories, and used them to create new regular expression patterns, obtaining a speculation-based version of NegEx. For the deep learning model, we implemented SpecB, finetuning a pretrained BERT model on 3190 BioScope sentences (280 of which with a speculation scope) for 10 epochs, learning rate 0.0001.

## Evaluation Metrics

We focus on two characteristics of the models: their ability to extract ADEs from the texts, and their robustness against negation and speculation.
To measure the first one, we use the standard metrics employed during the SMM4H shared tasks,[8] in particular the Relaxed Precision (P), Recall (R) and F1 score (F1):

$$\mathrm{R} = (\mathrm{TP} + 0.5 \times \mathrm{Par}) \ / \ (\mathrm{TP} + \mathrm{Par} + \mathrm{FN})$$

$$\mathrm{P} = (\mathrm{TP} + 0.5 \times \mathrm{Par}) \ / \ (\mathrm{TP} + \mathrm{Par} + \mathrm{FP})$$

$$\mathrm{F1} = (2 \times \mathrm{P} \times \mathrm{R}) \ / \ (\mathrm{P} + \mathrm{R})$$

These scores also take into account partial matches ($\mathrm{Par}$), and thus it is sufficient for a prediction to partially overlap with the real annotation. Figure 4 illustrates the meaning of True Positive (TP), False Negative (FN), False Positive (FP) and Partial matches in this task.
To explicitly measure the robustness of the models against the two linguistic phenomena, we also keep track of the number of FP on the various test set partitions. For example, a high number of FP on the $\mathbb{N}$ samples shows that the model is not robust against negations, while the presence of FP among the $\mathbb{S}$ samples shows a poor understanding of speculations.

Figure 4. Illustration of the possible matches between predicted and real ADEs.

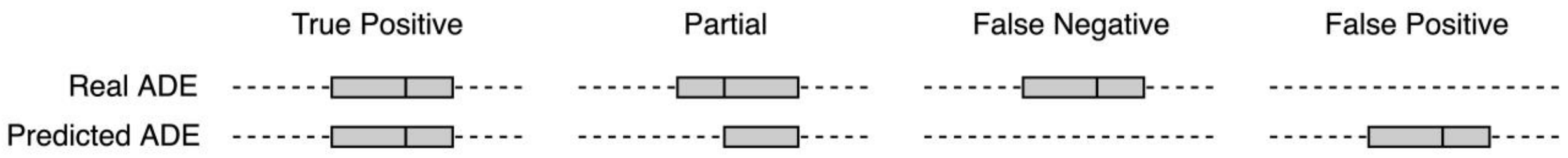

## Experiments

All experiments are carried out using the SNAX dataset to analyze the robustness of current ADE extraction systems on negations and speculations. The dataset is divided into a train and test set (see Table 1). All test samples consist of real tweets, while the training samples are a mix of real ($\mathbb{A}$ and $\mathbb{X}$) and artificially altered ones ($\mathbb{N}$ and $\mathbb{S}$). All the models are trained on the $\mathbb{A}$ and $\mathbb{X}$ samples, and tested on all data categories.

Table 1. Distribution of $\mathbb{A}$, $\mathbb{X}$, $\mathbb{N}$ and $\mathbb{S}$ samples in the SNAX train and test set. The $\mathbb{N}$ and $\mathbb{S}$ samples in the test set are all real samples, while the ones in the training set were generated artificially by the annotators.

| | **S** | **N** | **A** | **X** | **Total** | **S(%)** | **N(%)** | **A(%)** | **X(%)** |
|---|---|---|---|---|---|---|---|---|---|
| **Train** | 227 | 251 | 846 | 778 | 2102 | 10.80 | 11.94 | 40.25 | 37.01 |
| **Test** | 73 | 73 | 200 | 194 | 540 | 13.52 | 13.52 | 37.04 | 35.93 |

We first analyze the performance of the four ADE extraction models (see section “ADE Extraction Models”) to understand their initial robustness and compare it with the following experiments.
We then test the efficacy of our two proposed strategies: data augmentation, adding to the training set the artificially generated ℕ and/or 𝕊 samples; and model combination, combining each of the ADE extraction models with a negation detection module (NegEx, NegB) and/or a speculation detection module (SpecEx, SpecB).
To conclude the experiments, we also investigate the combined use of the two robustness-increasing strategies, combining the negation/speculation models with the base ADE models trained on the augmented dataset. To provide statistically significant results, all the reported metrics are averaged over 5 runs. During the analyses, particular attention has been paid to the number of false positive entities (FP) extracted by the models, because this metric clearly shows how many ADEs have been erroneously extracted due to the model’s lack of understanding negations and speculations.

# Results

To evaluate the performance of the ADE extract models (BERT, SpanBERT, PubMedBERT and EnDR-BERT) on their own, we train them on the 𝔸 and 𝕏 samples, and test them on all the available sample categories. The Table 2 shows that all the models have a similar FP distribution: most of the FP belong to the ℕ category, showing that the models have trouble understanding the effect of negations. The 𝕊 and 𝕏 categories tend to have a similar number of FP.

Table 2. Performance of the three baseline models on the test set of SNAX. The performance is measured in terms of False Positives (FP), Precision (P), Recall (R) and F1 score (F1).

| Extra train samp. | Model | FP | | | | | P | R | F1 |
|---|---|---|---|---|---|---|---|---|---|
| | | Total | S | N | A | X | | | |
| | BERT | 209.8 | 41.8 | 74.0 | 50.4 | 43.6 | 44.46 | 65.68 | 53.01 |
| | SpanBERT | 270.0 | 56.7 | 90.3 | 51.3 | 71.7 | 42.18 | 75.25 | 54.05 |
| | PubMedBERT | 196.6 | 44.4 | 65.6 | 39.0 | 47.6 | 47.81 | 70.21 | 56.85 |
| | EnDR-BERT | 189.8 | 46.2 | 60.6 | 38.0 | 45.0 | 49.87 | 72.86 | 59.21 |

In the second experiment, we apply the data augmentation strategy, adding the artificial ℕ and/or 𝕊 samples to the training set. Looking at Table 3, we can see that adding ℕ samples in the training set strongly reduces the number of FP in the ℕ category with respect to the baseline models (e.g., from 74.0 to 29.6 for BERT and from 60.6 to 15.4 for EnDR-BERT), while also helping to lower the number of FP in the other categories. Similarly, adding 𝕊 samples set strongly reduces the number of FP in the 𝕊 category (e.g., from 41.8 to 28.6 for BERT and from 46.2 to 16.2 for EnDR-BERT) and 𝕏 category, helping to lower the number of FP in the other categories.

In both cases F1 and P increase by 2–5 points, despite a loss of up to 5 points in R. R decreases because the models are now more likely to discard some actual ADE, but the sharp increase in P shows that when they detect an ADE, it is more likely to be an actual one.

Adding all generated samples to the training set (ℕ+𝕊) compounds the previous effects, lowering the number of FP in all categories and contributing to a further increase in P and F1. For example, if we compare EnDR-BERT with ℕ+𝕊 EnDR-BERT (Table 3), R decreases by 4.45 points, while P increases by 11.80 points, leading to an overall increase in F1 from 59.21 to 64.83.

The tradeoff between P (low False Positive rate) and R (high False Negatives rate) was mostly expected. As stated in the Introduction, in the context of social listening with great amounts of data, noise reduction (high Precision) is more crucial than keeping a high Recall, as the increase of False Negatives does not have a direct negative impact on the downstream task. However, the drop in R for this strategy is minimal, and could therefore be a good candidate even for situations where high R is a requirement.

Table 3. Data augmentation strategy: performance of the three base models trained on the augmented dataset, with the addition of ℕ and/or 𝕊 samples. The performance is measured in terms of False Positives (FP), Precision (P), Recall (R) and F1 score (F1).

| **Extra train samp.** | **Model** | **FP** | | | | | **P** | **R** | **F1** |
|---|---|---|---|---|---|---|---|---|---|
| | | **Total** | **S** | **N** | **A** | **X** | | | |
| | BERT | 209.8 | 41.8 | 74.0 | 50.4 | 43.6 | 44.46 | **65.68** | 53.01 |
| N | BERT | 133.8 | 36.4 | 29.6 | **37.2** | 30.6 | 51.92 | 60.15 | 55.71 |
| S | BERT | 153.0 | 28.6 | 56.6 | 39.8 | 28.0 | 50.26 | 62.28 | 55.60 |
| N+S | BERT | **107.8** | **26.6** | **21.4** | 38.2 | **21.6** | **56.33** | 60.21 | **58.13** |
| | SpanBERT | 270.0 | 56.7 | 90.3 | 51.3 | 71.7 | 42.18 | **75.25** | 54.05 |
| N | SpanBERT | **181.5** | 45.0 | **44.8** | **41.8** | 50.0 | **49.74** | 71.14 | **58.51** |
| S | SpanBERT | 197.0 | **30.0** | 76.0 | 49.0 | **42.0** | 48.11 | 71.06 | 57.30 |
| N+S | SpanBERT | 198.5 | 40.8 | 60.5 | 50.5 | 46.8 | 47.36 | 68.73 | 55.84 |
| | PubMedBERT | 196.6 | 44.4 | 65.6 | 39.0 | 47.6 | 47.81 | **70.21** | 56.85 |
| N | PubMedBERT | 117.0 | 34.6 | 20.6 | 28.8 | 33.0 | 56.15 | 62.00 | 58.78 |
| S | PubMedBERT | 142.4 | **14.2** | 58.8 | 37.6 | 31.8 | 53.14 | 65.32 | 58.56 |
| N+S | PubMedBERT | **91.2** | 14.4 | **25.6** | **27.2** | **24.0** | **59.92** | 59.45 | **59.84** |
| | EnDR-BERT | 189.8 | 46.2 | 60.6 | 38.0 | 45.0 | 49.87 | **72.86** | 59.21 |
| N | EnDR-BERT | 119.4 | 41.0 | **15.2** | **29.6** | 33.6 | 58.02 | 68.69 | 62.90 |
| S | EnDR-BERT | 133.0 | **16.2** | 51.6 | 35.6 | **29.6** | 56.85 | 70.55 | 62.93 |
| N+S | EnDR-BERT | **99.0** | 22.0 | 15.4 | 31.6 | 30.0 | **61.67** | 68.41 | **64.83** |

In the third experiment, we apply the model combination strategy, combining each ADE extraction model with a negation or speculation extraction module (NegEx, NegB, SpecEx or SpecB).

Table 4 shows that adding a negation detection module only lowers the number of FP in the ℕ category, without impacting the others. Adding a speculation detection module, on the other hand, strongly lowers the number of FP in the 𝕊 category (e.g., from 44.4 for PubMedBERT to 8.4 for PubMedBERT°SpecEx), and also affects the 𝕏 category. This again shows that the performances on the 𝕊 and 𝕏 categories seem to be highly correlated. The drop in FP is way sharper that the one obtained with the data augmentation strategy, adding the 𝕊 samples to the training set. However, the drop in R is so strong that it also affects the F1 score, lowering it by 3 points.
Finally, we add a module for each of the linguistic phenomena, choosing NegB and SpecB for their higher R, and apply both of them to the ADE extraction models (NegB°SpecB). The effects previously described for the single modules both apply, without interfering with each other, leading to an overall lower number of FP and a drop in F1 score.

Table 4. Model combination strategy: performance of the models obtained combining the base ADE extraction models with a negation and/or speculation detection module. The performance is measured in terms of False Positives (FP), Precision (P), Recall (R) and F1 score (F1).

| Extra train samp. | Model | FP | | | | | P | R | F1 |
|---|---|---|---|---|---|---|---|---|---|
| | | Total | S | N | A | X | | | |
| | BERT | 209.8 | 41.8 | 74.0 | 50.4 | 43.6 | 44.46 | **65.68** | 53.01 |
| | BERT°NegEx | 153.0 | 38.6 | **26.4** | 45.6 | 42.4 | 48.54 | 58.62 | **53.09** |
| | BERT°NegB | 163.8 | 40.6 | 37.6 | 42.8 | 42.8 | 47.61 | 59.90 | 53.04 |
| | BERT°SpecEx | 125.2 | 8.2 | 43.8 | 41.6 | 31.6 | 49.34 | 50.54 | 49.90 |
| | BERT°SpecB | 136.4 | **7.8** | 60.0 | 41.8 | 26.8 | 48.95 | 53.54 | 51.12 |
| | BERT°NegB°SpecB | **101.4** | **7.8** | 33.4 | **34.2** | **26.0** | **52.54** | 48.17 | 50.25 |
| | SpanBERT | 270.0 | 56.7 | 90.3 | 51.3 | 71.7 | 42.18 | **75.25** | 54.05 |
| | SpanBERT°NegEx | 196.0 | 51.0 | **32.7** | 43.7 | 68.7 | 46.10 | 65.74 | 54.18 |
| | SpanBERT°NegB | 206.0 | 52.0 | 43.7 | 43.0 | 67.3 | 45.96 | 68.58 | **55.03** |
| | SpanBERT°SpecEx | 156.0 | 13.0 | 51.0 | 39.0 | 53.0 | 47.81 | 55.96 | 51.43 |
| | SpanBERT°SpecB | 173.7 | 12.3 | 70.7 | 44.0 | 46.7 | 47.09 | 60.28 | 52.87 |
| | SpanBERT°NegB°SpecB | **128.3** | **11.7** | 38.3 | **35.7** | **42.7** | **51.21** | 54.20 | 52.66 |
| | PubMedBERT | 196.6 | 44.4 | 65.6 | 39.0 | 47.6 | 47.81 | **70.21** | 56.85 |
| | PubMedBERT°NegEx | 143.6 | 38.8 | **24.4** | 34.8 | 45.6 | 51.89 | 62.29 | 56.58 |
| | PubMedBERT°NegB | 153.6 | 41.4 | 33.0 | 32.4 | 46.8 | 51.13 | 64.56 | **57.04** |
| | PubMedBERT°SpecEx | 115.6 | **8.4** | 37.4 | 32.4 | 37.4 | 52.66 | 52.87 | 52.73 |
| | PubMedBERT°SpecB | 120.2 | 8.6 | 49.2 | 33.6 | 28.8 | 53.50 | 56.26 | 54.81 |
| | PubMedBERT°NegB°SpecB | **91.2** | **8.4** | 27.8 | **27.0** | **28.0** | **56.92** | 51.11 | 53.83 |
| | EnDR-BERT | 189.8 | 46.2 | 60.6 | 38.0 | 45.0 | 49.87 | **72.86** | 59.21 |
| | EnDR-BERT°NegEx | 137.2 | 42.6 | **15.8** | 35.0 | 43.8 | 54.37 | 65.36 | **59.36** |
| | EnDR-BERT°NegB | 151.4 | 45.0 | 28.0 | 33.4 | 45.0 | 52.94 | 67.24 | 59.24 |
| | EnDR-BERT°SpecEx | 107.2 | 8.8 | 32.0 | 30.2 | 36.2 | 56.51 | 55.93 | 56.21 |
| | EnDR-BERT°SpecB | 109.0 | **5.8** | 44.8 | 32.4 | **26.0** | **57.32** | 59.48 | 58.37 |
| | EnDR-BERT°NegB°SpecB | **82.2** | **5.8** | 22.6 | **27.8** | 26.0 | 60.81 | 53.98 | 57.19 |

Finally, we apply both strategies at the same time to see if they combine in a positive way. We add all the artificial samples to the training dataset and combine the ADE extraction models with both NegB and SpecB. The two strategies are complementary, in the sense that the number of FP decreases further, when compared to using just one of them (see Table 5). For example, the total number of FP for EnDR-BERT was 99.0 when using the data augmentation strategy, 107.2 when using the model combination strategy and 53.6 when using both. In general, these models have about ¼ of the total FPs generated by the baseline one (gray rows). However, this has a very negative impact on the R, which decreases on average by 20 points. This is counterbalanced by an increase in P (15 points on average), which leads to these models having roughly the same F1 score as the baseline ones.

Table 5. Data augmentation and model combination strategy: performance of the models obtained combining the base ADE extraction models with a negation and speculation detection module, and training them on the augmented dataset. The performance is measured in terms of False Positives (FP), Precision (P), Recall (R) and F1 score (F1).

| Extra train samp. | Model | FP | | | | | P | R | F1 |
|---|---|---|---|---|---|---|---|---|---|
| | | Total | S | N | A | X | | | |
| | BERT | 209.8 | 41.8 | 74.0 | 50.4 | 43.6 | 44.46 | **65.68** | **53.01** |
| N+S | BERT°NegB°SpecB | **65.8** | **7.2** | **12.0** | **31.0** | **15.6** | **59.88** | 45.76 | 51.82 |
| | SpanBERT | 270.0 | 56.7 | 90.3 | 51.3 | 71.7 | 42.18 | **75.25** | **54.05** |
| N+S | SpanBERT°NegB°SpecB | **106.8** | **10.0** | **26.5** | **38.8** | **31.5** | **54.16** | 51.13 | 52.42 |
| | PubMedBERT | 196.6 | 44.4 | 65.6 | 39.0 | 47.6 | 47.81 | **70.21** | **56.85** |
| N+S | PubMedBERT°NegB°SpecB | **56.0** | **5.4** | **13.4** | **21.8** | **15.4** | **64.14** | 46.38 | 53.73 |
| | EnDR-BERT | 189.8 | 46.2 | 60.6 | 38.0 | 45.0 | 49.87 | **72.86** | **59.21** |
| N+S | EnDR-BERT°NegB°SpecB | **53.6** | **3.4** | **5.8** | **23.8** | **20.6** | **67.66** | 52.37 | 59.00 |

## Take-home messages

As highlighted by the analyses above, both proposed strategies have an important impact on the models' performances, and the effect is the same regardless of the chosen base model. For reference, Figure 5 offers a visual comparison of all the metrics for one of the ADE models (EnDR-BERT).

Figure 5. Visual comparison of the metrics for all experiments on EnDR-BERT. Applying both strategies leads to the lowest number of FP in all categories, but using data augmentation alone offers a better balance of FP and R.

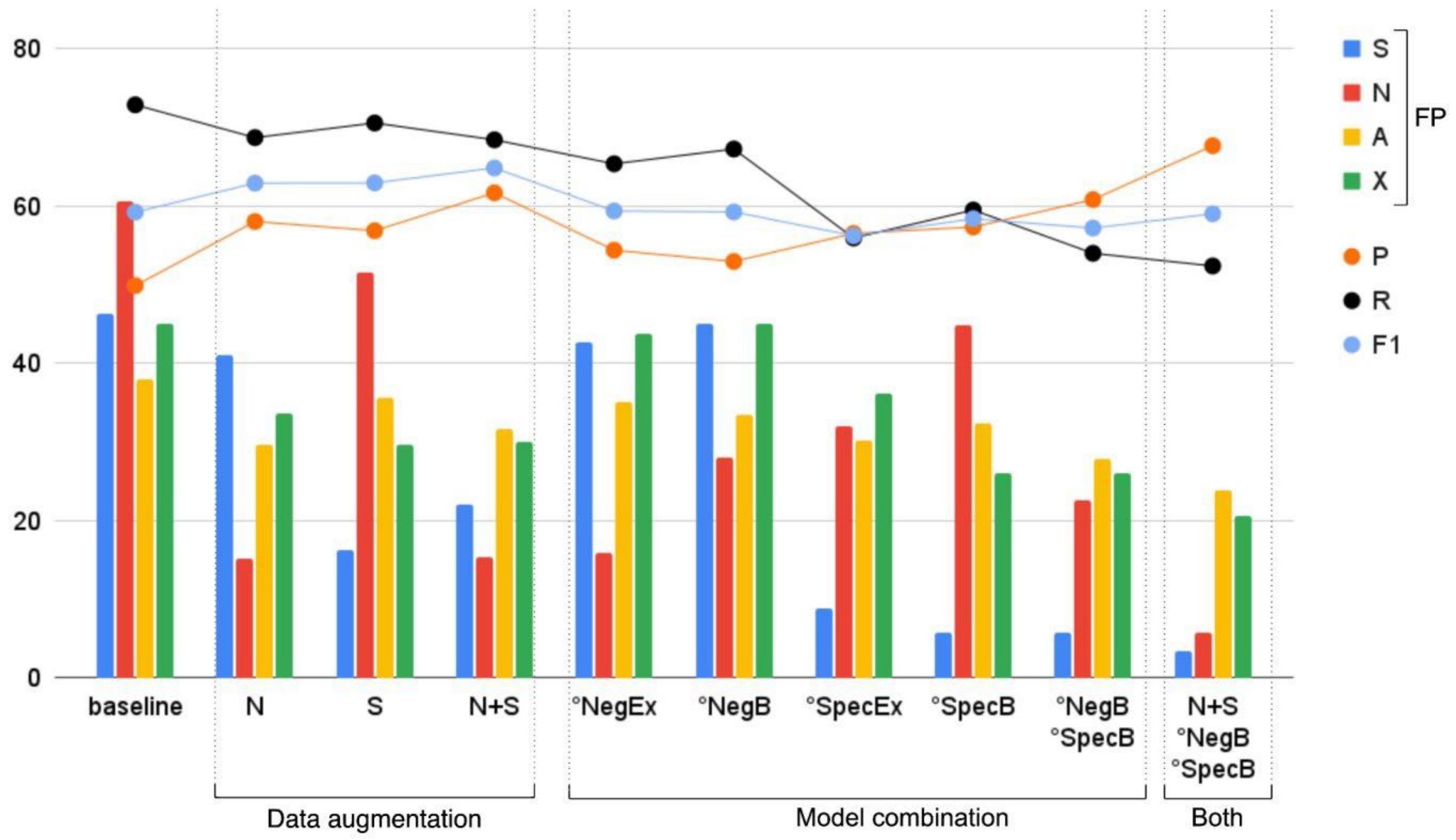

We summarize our main findings and observations as follows:

- All the baseline ADE detection models that we analyzed are not robust against negations and speculations. Negations seem to be a more difficult category for models which are not specifically designed to deal with them, while the effect of speculations seems similar to that of the general samples that do not contain ADEs ($\mathbb{X}$ category).

- The two strategies that we introduce, both lower the number of FP predictions of the baselines models and affect all models in the same way. They can therefore be applied to future BERT-based models, having clear expectations of their effects.

- Introducing a small number of new samples (data augmentation strategy), even if artificial, seems to be the best way to directly increase the model's knowledge about the phenomena, without excessively impacting the Recall of the base model.

- In cases where adding new training data is not a feasible strategy, combining the baseline ADE model with a negation/speculation detection model (model combination strategy) reliably leads to a reduction in the number of FPs (only in the specific negation/speculation category), but can lower the Recall of the models significantly.

- Regex-based modules and deep-learning-based modules give similar results when applied in the model combination strategy, so a well-designed regex module can be sufficient in absence of large amounts of data to train specific deep-learning models for the linguistic phenomenon of interest.

- Compounding the data augmentation strategy for negation and speculations (i.e., adding both kinds of samples to the training data) has only positive effects on all of the monitored metrics, and it is therefore encouraged.

- Compounding the model combination strategy for negation and speculations (i.e., adding both a negation and a speculation detection module) has a positive effect on the FP reduction, but negative effects on Recall and F1 score, and it is therefore discouraged.

- Applying both strategies leads to the best results in terms of FP reduction, but exacerbates the collateral effects of the model combination strategy. It is therefore highly discouraged to use both strategies if interested in keeping a high Recall. On the other hand, this is the best way to achieve a 15-point increase in Precision.

# Discussion

In this paper, we evaluated the impact of negations and speculations on state-of-the-art ADE detection models. We introduced SNAX, a new dataset specifically aimed at studying these phenomena. The dataset proves to be a challenging setting and the experiments show that current methods lack mechanisms to deal with negations and speculations.

We introduce and compare two strategies to tackle the problem: adding artificially negated and speculated samples in the training set, and combining the ADE extraction model with a specialized negation and/or speculation detection module. Both of them bring significant increases in performance. Both the dataset and the code are made publicly available for the community to test the robustness of their systems against negations and speculations.

Some interesting directions for future works would be: investigating other ways to merge the predictions of ADE extraction and negation/speculation detection modules (e.g., with a joint learning approach), and increasing the quality and quantity of real negated and speculated samples via crowd-sourcing initiatives or multi-round adversarial benchmarking.

# Author's Contributions

All authors participated in the design, interpretation of the studies, analysis of the data, and review of the manuscript. SS and BP conducted the experiments; SS and EC performed quality control on the generated samples; GS, EC, and ES recruited the annotators and supervised the experimental process.

# Declaration of Conflicting Interests

Enrico Santus is a Senior Lead Data Scientist at Bayer Pharmaceuticals.

# Funding

This research received no specific grant from any funding agency in the public, commercial, or not-for-profit sectors.

# References

1. Alvarez-Requejo A, Carvajal A, Bégaud B, Moride Y, Vega T, Martín Arias LH. Under-reporting of adverse drug reactions. Estimate based on a spontaneous reporting scheme and a sentinel system. *Eur J Clin Pharmacol* 1998;**54**:483-8
2. Karimi S, Wang C, Metke-Jimenez A, Gaire R, Paris C. Text and data mining techniques in adverse drug reaction detection. *ACM Comput Surv* 2015;**47**:1-39
3. Sarker A, Gonzalez G. Portable automatic text classification for adverse drug reaction detection via multi-corpus training. *J Biomed Inform* 2015;**53**:196-207
4. Sarker A, Ginn R, Nikfarjam A, O'Connor K, Smith K, Jayaraman S, Upadhaya T, Gonzalez G. *J Biomed Inform* 2015;**54**:202-12
5. Paul MJ, Sarker A, Brownstein JS, Nikfarjam A, Scotch M, Smith KL, Gonzalez G. Social media mining for public health monitoring and surveillance. *Pacific Symp Biocomp* 2016;468-79
6. Sarker A, Gonzalez-Hernandez G. Overview of the second social media mining for health (SMM4H) shared tasks at AMIA 2017. *CEUR Workshop Proc* 2017;43-8
7. Weissenbacher D, Sarker A, Paul MJ, Gonzalez-Hernandez G. Overview of the Third Social Media Mining for Health (SMM4H) Shared Tasks at EMNLP 2018. *Proc 2018 EMNLP Work SMM4H: 3rd Soc Media Min Heal Appl Work Shar Task* 2018;13-6
8. Weissenbacher D, Sarker A, Magge A, Daughton A, O'Connor K, Paul MJ, Gonzalez-Hernandez G. Overview of the Fourth Social Media Mining for Health (SMM4H) Shared Tasks at ACL 2019. *Proc Fourth Soc Media Min Heal Appl (#SMM4H) Work Shar Task* 2019;21-30
9. Klein A, Alimova I, Flores I, Magge A, Miftahutdinov Z, Minard A-L, O'Connor K, Sarker A, Tutubalina E, Weissenbacher D, Gonzalez-Hernandez G. Overview of the Fifth Social Media Mining for Health Applications (#SMM4H) Shared Tasks at COLING 2020. *Proc Fifth Soc Media Min Heal Appl Work Shar Task* 2020;27-36
10. Vaswani A, Brain G, Shazeer N, Parmar N, Uszkoreit J, Jones L, Gomez AN, Kaiser Ł, Polosukhin I. Attention Is All You Need. *Adv Neural Inf Process Syst* 2017;**30**:6000-10
11. Devlin J, Chang MW, Lee K, Toutanova K. BERT: Pre-training of deep bidirectional transformers for language understanding. *NAACL HLT 2019* 2019;**1**: 4171-86
12. Velldal E, Øvrelid L, Read J, Oepen S. Speculation and Negation: Rules, Rankers, and the Role of Syntax. *Comput Linguistics* 2012;**38**:369-410
13. Cruz Díaz NP. Detecting negated and uncertain information in biomedical and review texts. *Proc Student Research Work Assoc Ranlp 2013* 2013;45-50
14. Scaboro S, Portelli B, Chersoni E, Santus E, Giuseppe S. NADE: A Benchmark for Robust Adverse Drug Events Extraction in Face of Negations. *Proc Seventh Work Noisy User-Generated Text (W-Nut 2021)* 2021;230-7

15. Nikfarjam A, Sarker A, O'Connor K, Ginn R, Gonzalez G. Pharmacovigilance from social media: Mining adverse drug reaction mentions using sequence labeling with word embedding cluster features. *J Am Med Inform Assoc* 2015;**22**:671-81
16. Daniulaityte R, Chen L, Lamy FR, Carlson RG, Thirunarayan K, Sheth A. "When 'bad' is 'good'": Identifying personal communication and sentiment in drug-related tweets. *JMIR Public Health Surveill* 2016;**2**:e162
17. Chen S, Huang Y, Huang X, Qin H, Yan J, Tang B. HITSZ-ICRC: A Report for SMM4H Shared Task 2019-Automatic Classification and Extraction of Adverse Effect Mentions in Tweets. *Proc Fourth Soc Media Min Heal Appl (#SMM4H) Work Shar Task* 2019;47-51
18. Mahata D, Anand S, Zhang H, Shahid S, Mehnaz L, Kumar Y, Shah RR. MIDAS@SMM4H-2019: Identifying Adverse Drug Reactions and Personal Health Experience Mentions from Twitter. *Proc Fourth Soc Media Min Heal Appl (#SMM4H) Work Shar Task* 2019.127-132
19. Miftahutdinov Z, Alimova I, Tutubalina E. KFU NLP Team at SMM4H 2019 Tasks: Want to Extract Adverse Drugs Reactions from Tweets? BERT to The Rescue. *Proc Fourth Soc Media Min Heal Appl (#SMM4H) Work Shar Task* 2019;52-57
20. Portelli B, Lenzi E, Chersoni E, Serra G, Santus E. BERT Prescriptions to avoid unwanted headaches: A comparison of transformer architectures for adverse drug event detection. *Proc 16th Conf Eur Chapter Assoc Comput Linguistics: Main Vol* 2021;1470-47
21. Ribeiro MT, Wu T, Guestrin C, Singh S. Beyond Accuracy: Behavioral Testing of NLP Models with CheckList. *Proc 58Th Annu Meeting Assoc Comput Linguistics* 2020;4902-12
22. Karimi S, Metke-Jimenez A, Kemp M, Wang C. Cadec: A corpus of adverse drug event annotations. *J Biomed Inform* 2015;**55**:73-81
23. Alvaro N, Miyao Y, Collier N. TwiMed: Twitter and PubMed comparable corpus of drugs, diseases, symptoms, and their relations. *JMIR Public Health Surveill* 2017;**3**
24. Chapman WW, Bridewell W, Hanbury P, Cooper GF, Buchanan BG. A simple algorithm for identifying negated findings and diseases in discharge summaries. *J Biomed Inform* 2001;**34**:301-10
25. Harkema H, Dowling JN, Thornblade T, Chapman WW. ConText: An algorithm for determining negation, experiencer, and temporal status from clinical reports. *J Biomed Inform* 2009;**42**:839-51
26. Chapman WW, Hillert D, Velupillai S, Kvist M, Skeppstedt M, Chapman BE, Conway M, Tharp M, Mowery DL, Deleger L. Extending the NegEx lexicon for multiple languages. *Stud Health Technol Inform* 2013;**192**:677-81
27. Vincze V, Szarvas G, Farkas R, Móra G, Csirik J. The BioScope corpus: Biomedical texts annotated for uncertainty, negation and their scopes. *BMC Bioinformatics* 2008;**9**:38-45
28. Morante R, Liekens A, Daelemans W. Learning the scope of negation in biomedical texts. *Proc 2008 Conf Empir Methods Natural Language Process* 2008;715-24.
29. Cruz Díaz NP, Maña López MJ, Vázquez JM, Álvarez VP. A machine-learning approach to negation and speculation detection in clinical texts. *J Am Soc Inf Sci Technol* 2012;**63**:1398-410
30. Attardi G, Cozza V, Sartiano D. Detecting the scope of negations in clinical notes. *Proc Second Italian Conf Comput Linguistics CLiC-It 2015* 2015;130-5
31. Zou B, Zhu Q, Zhou G. Negation and speculation identification in Chinese language. *Proc 53rd Annu Meeting Assoc Comput Linguistics 7th Int Jt Conf Natural Language Process* 2015;**1**:656-65

32. Qian Z, Li P, Zhu Q, Zhou G, Luo Z, Luo W. Speculation and negation scope detection via convolutional neural networks. *Proc 2016 Conf Empir Methods Natural Language Process* 2016;815-25
33. Fancellu F, Lopez A, Webber B. Neural networks for negation scope detection. *Proc 54th Annu Meeting Assoc Comput Linguistics* 2016;**1**:495-504
34. Fancellu F, Lopez A, Webber B, He H. Detecting negation scope is easy, except when it isn't. *Proc 15th Conf Eur Chapter Assoc Comput Linguistics* 2017;**2**:58-63
35. Dalloux C, Claveau V, Grabar N. Speculation and negation detection in French biomedical corpora. *Proc Int Conf Recent Adv Natural Language Process (RANLP 2019)* 2019;223-32
36. Khandelwal A, Sawant S. NegBERT: A transfer learning approach for negation detection and scope resolution. *Proc 12th Language Resour Evaluation Conf* 2020;5739–48
37. Zavala RR, Martinez P. The impact of pretrained language models on negation and speculation detection in cross-lingual medical text: Comparative study. *JMIR Med Inform* 2020;**8**:e18953
38. Khandelwal A, Britto BK. Multitask learning of negation and speculation using transformers. *Proc 11th Int Work Heal Text Min Inf Anal* 2020;79-87
39. Kiela D, Bartolo M, Nie Y, Kaushik D, Geiger A, Wu Z, Vidgen B, Prasad G, Singh A, Ringshia P, Ma Z, Thrush T, Riedel S, Waseem Z, Stenetorp P, Jia R, Bansal M, Potts C, Williams A. Dynabench: Rethinking Benchmarking in NLP. *Proc 2021 Conf North Am Chapter Assoc Comput Linguistics: Hum Language Technol* 2021;4110-24
40. Potts C, Wu Z, Geiger A, Kiela D. Dynasent: A dynamic benchmark for sentiment analysis. *Proc 59th Annu Meeting Assoc Comput Linguistics 11th Int Jt Conf Natural Language Process* 2020;**1**:2388-404
41. Bartolo M, Thrush T, Jia R, Riedel S, Stenetorp P, Kiela D. Improving question answering model robustness with synthetic adversarial data generation. *Proc 2021 Conf Empir Methods Natural Language Process* 2021;8830-48
42. Vidgen B, Thrush T, Waseem Z, Kiela D. Learning from the worst: Dynamically generated datasets to improve online hate detection. *Proc 59th Annu Meeting Assoc Comput Linguistics 11th Int Jt Conf Natural Language Process* 2020;**1**:1667-82
43. Portelli B, Passabì D, Lenzi E, Serra G, Santus E, Chersoni E. Improving Adverse Drug Event Extraction with SpanBERT on Different Text Typologies. *AI Disease Surveillance Pandemic Intell. W3PHAI 2021. Stud Comput Intell* 2022;**1013**:87-99
44. Joshi M, Chen D, Liu Y, Weld DS, Zettlemoyer L, Levy O. Spanbert: Improving pre-training by representing and predicting spans. *Trans Assoc Comput Linguist* 2020;**8**:64-77
45. Gu Y, Tinn R, Cheng H, Lucas M, Usuyama N, Liu X, Naumann T, Gao J, Poon H. Domain-Specific Language Model Pretraining for Biomedical Natural Language Processing. *ACM Trans Comput Healthcare* 2022;**3**:1-23
46. Miftahutdinov Z, Sakhovskiy A, Tutubalina E. KFU NLP Team at SMM4H 2020 Tasks: Cross-lingual Transfer Learning with Pretrained Language Models for Drug Reactions. *Proc Fifth Soc Media Min Heal Appl Work Shar Task* 2020;51-6